\documentclass[runningheads]{llncs}

\usepackage[T1]{fontenc}
\usepackage{lmodern}
\usepackage{microtype}
\usepackage{graphicx}
\usepackage{booktabs}
\usepackage{tabularx}
\usepackage{array}
\usepackage{adjustbox}
\usepackage{multirow}
\usepackage{url}
\usepackage{amsmath}
\usepackage{amssymb}
\usepackage{placeins}
\usepackage{float}
\usepackage{xcolor}
\usepackage{capt-of}
\usepackage{tikz}
\usetikzlibrary{shapes,arrows.meta,positioning,fit,backgrounds,calc,shadows.blur}

\DeclareMathOperator*{\argmax}{arg\,max}

\newcommand{\figplaceholder}[2]{%
\IfFileExists{Figures/#1}{\includegraphics[width=#2]{Figures/#1}}{%
\fbox{\parbox[c][0.20\textheight][c]{#2}{\centering Missing figure: \\ \texttt{\detokenize{Figures/#1}}}}}%
}

\definecolor{tierData}{HTML}{D6E4F0}
\definecolor{tierRep}{HTML}{D9EAD3}
\definecolor{tierEval}{HTML}{FCE5CD}
\definecolor{tierMeta}{HTML}{EFEFEF}
\definecolor{borderData}{HTML}{4A6FA5}
\definecolor{borderRep}{HTML}{4F8A4A}
\definecolor{borderEval}{HTML}{C97A3F}
\definecolor{borderMeta}{HTML}{888888}

\usepackage{orcidlink}
\renewcommand{\orcidID}[1]{\,\orcidlink{#1}}

\begin{document}

\title{When Does Deep Representation Learning Help Single-Cell Clustering? A Sensitivity-Aware Diagnostic Benchmark for Biomedical AI Pipelines}
\titlerunning{When Does Deep Representation Learning Help scRNA-seq Clustering?}

\author{Nguyen Thanh Phong\inst{1}\orcidID{0009-0001-1028-4305
}, Truong Viet Vu\inst{1}\orcidID{0009-0004-3941-1306}, Nguyen Ha Thu\inst{1}\orcidID{0009-0005-7694-6497}, Tran An Ky\inst{1}\orcidID{0009-0002-1076-2693}, Tran Hoang Thong\inst{1}\orcidID{0009-0000-0453-5799}, Le Pham Thuy Hien\inst{2}\orcidID{0009-0008-9663-0352}, and
Nguyen Thai Anh\inst{1}\orcidID{0009-0005-5600-6510}}
\titlerunning{When Does Deep Representation Learning Help scRNA-seq Clustering?}
\authorrunning{Phong et al.}
\institute{
Faculty of Information Technology, Van Lang School of Technology, Van Lang University, Ho Chi Minh City, Vietnam \and
Faculty of Law, Van Lang University, Ho Chi Minh City, Vietnam\\
\email{anh.nt@vlu.edu.vn}}

\maketitle

\begin{abstract}
Single-cell ribonucleic acid sequencing (scRNA-seq) is a foundational technology for precision-medicine workflows that contribute to United Nations Sustainable Development Goal~3 on Good Health and Well-being, and unsupervised clustering is the analytical step that turns raw expression matrices into interpretable cell populations. Practitioners therefore face a recurring engineering decision: is an additional deep representation stage worth its compute and tuning cost, or do classical principal component analysis (PCA) pipelines already suffice? We address this question with a diagnostic benchmark of nine clustering pipelines on ten real datasets ($90$--$5{,}685$ cells, $19{,}046$--$41{,}480$ genes, $4$--$11$ cell types), augmented by a partial scVI~V2 specialized comparison on seven datasets. The protocol integrates Optuna hyperparameter search, repeated-run robustness, Friedman/Wilcoxon-Holm/TOST testing, and Sobol total-order sensitivity analysis. The contrastive autoencoder achieved the highest mean Adjusted Rand Index ($0.7872$), but Holm-corrected tests did not establish dominance over the strongest baselines. Per-dataset analysis reveals three reproducible regimes: probabilistic variational autoencoder (VAE) variants help on the smallest datasets, deep autoencoders win on mid-scale data with multi-batch or many-type structure, and classical PCA pipelines remain competitive when linear projection already captures the dominant variation. Sobol indices identify learning rate ($S_T{=}0.70$) and latent dimensionality ($S_T{=}0.56$) as the dominant variance contributors, indicating where limited tuning budgets should be allocated. The contribution is therefore a dataset-aware and compute-conscious decision framework for biomedical AI pipelines supporting sustainable healthcare analytics, rather than a universal superiority claim.

\end{abstract}

\section{Introduction}
Artificial intelligence (AI) is reshaping biomedical research and clinical practice, supporting United Nations Sustainable Development Goal~3 on Good Health and Well-being. Single-cell ribonucleic acid sequencing (scRNA-seq) profiles gene expression at individual-cell resolution and has become a standard technology for studying cellular heterogeneity, rare cell populations, developmental processes, and disease-associated states~\cite{tang2009mrna,kolodziejczyk2015technology,luecken2019current}. Reliable analysis of scRNA-seq matrices underpins precision-medicine applications such as cell-type discovery, drug-target identification, and patient stratification~\cite{kiselev2019challenges}. Yet the data are high-dimensional, sparse, noisy, and affected by donor or batch variation~\cite{kiselev2019challenges,korsunsky2019fast}, and the standard analytical step, unsupervised clustering, remains sensitive to preprocessing, representation quality, and clustering hyperparameters.

\paragraph{Existing pipelines.} Classical scRNA-seq workflows are commonly built on the Scanpy ecosystem~\cite{wolf2018scanpy}, where raw count-like matrices are filtered, normalized, transformed, and reduced before clustering. Typical pipelines combine highly variable gene selection, principal component analysis (PCA), neighborhood graph construction, uniform manifold approximation and projection (UMAP) visualization~\cite{mcinnes2018umap}, and clustering through KMeans, Gaussian mixture models (GMM), Leiden~\cite{traag2019louvain}, or HDBSCAN~\cite{mcinnes2017hdbscan,campello2013density}. When batch or donor annotations are available, integration methods such as Harmony reduce technical variation in the learned embedding~\cite{korsunsky2019fast}. Deep representation learning provides an alternative through autoencoders~\cite{hinton2006reducing}, variational autoencoders (VAEs)~\cite{kingma2013auto}, or specialized single-cell generative models such as scVI~\cite{lopez2018deep}, and model-based deep embedded clustering methods such as scDeepCluster~\cite{tian2019clustering}. However, deep models are not automatically superior; their performance depends on input representation, latent dimensionality, objective function, batch metadata, and the downstream clustering head~\cite{lopez2018deep}, so controlled benchmarking remains necessary.

\paragraph{Position and motivation.} Existing studies and tutorials often examine one component of the pipeline in isolation. We instead pose a diagnostic question: \emph{under what dataset conditions does an additional deep representation stage produce a measurable benefit, and which hyperparameters drive most of the variance?} Answering this matters in practice. Trial-and-error tuning consumes compute that could be redirected to other stages of the analysis, and quantifying which decisions actually move the metric allows tighter budgets in real biomedical projects. This view aligns clustering practice with broader goals of compute-efficient and trustworthy biomedical AI.

\paragraph{Contributions.} (i)~A ten-dataset diagnostic benchmark with explicit metadata profiling, comparing six classical baselines, three deep autoencoder variants, and a partial scVI~V2 specialized baseline. (ii)~A statistical protocol combining Optuna search~\cite{akiba2019optuna}, repeated-run robustness, Friedman/Wilcoxon-Holm/TOST testing~\cite{friedman1937use,wilcoxon1992individual,holm1979simple,schuirmann1987comparison}, and Sobol total-order sensitivity~\cite{sobol1990sensitivity,saltelli2010variance}. (iii)~Three dataset-regime rules derived from per-dataset winners, supporting dataset-aware method selection rather than a one-size-fits-all recommendation. (iv)~A hyperparameter-importance ranking that identifies learning rate and latent dimensionality as the dominant variance contributors in our deep representation workflow. Overall, this study positions scRNA-seq clustering as part of a dataset-aware and compute-conscious biomedical AI pipeline for sustainable healthcare analytics.

\paragraph{Architecture naming.} The proposed architecture family is \mbox{DeepAE--UMAP--HDBSCAN}, with two variants: \mbox{C-DeepAE--UH} (contrastive) and \mbox{VAE--UH} (variational). The suffix ``UH'' refers to the UMAP--HDBSCAN clustering head. We include scVI~V2 as a partial specialized comparison; it completed on seven of ten datasets.

\section{Methodology}
\subsection{Problem Definition and Notation}
Let $X\in\mathbb{R}^{n\times p}$ denote the input expression matrix with $n$ cells and $p$ genes. We learn a representation $f_{\theta}:X\rightarrow Z$ with $Z\in\mathbb{R}^{n\times d}$ and $d\ll p$, where row $z_i$ encodes cell $i$ in the latent space. A clustering function then assigns predicted labels $\hat{y}_i=g(z_i)\in\{1,\ldots,K\}$. Reference annotations $y_i$ are used only for evaluation and for explicitly reported oracle-$k$ heads; they are never used to fit the unsupervised representation models.

\subsection{Preprocessing and Batch-Aware Correction}
Cells and genes are filtered to remove low-information entries, highly variable genes are selected, and the resulting matrix is scaled or projected depending on the downstream method. For datasets with donor or batch metadata (five of ten), a batch-aware step using the available batch key reduces technical variation in the representation stage; otherwise the same workflow runs without batch correction. This distinction matters because mixing the two regimes without acknowledgement would confound any representation comparison.

\subsection{Autoencoder, Contrastive AE, and VAE Objectives}
For the autoencoder (AE), the encoder $f_{\theta}$ maps $x_i$ to $z_i=f_{\theta}(x_i)$ and the decoder $h_{\phi}$ reconstructs $\hat{x}_i=h_{\phi}(z_i)$. The AE loss \cite{kunin2019loss} is
\begin{equation}
\mathcal{L}_{\mathrm{AE}}
=
\frac{1}{n}\sum_{i=1}^{n}
\ell\!\left(x_i, h_{\phi}(f_{\theta}(x_i))\right)
+ \lambda \|(\theta,\phi)\|_2^2 ,
\end{equation}
where $\ell$ is the reconstruction loss and $\lambda$ controls $L_2$ regularization. The contrastive AE (CAE) adds a regularization term $\mathcal{L}_{\mathrm{CAE}}=\mathcal{L}_{\mathrm{AE}}+\alpha\mathcal{L}_{\mathrm{contrast}}$. Specifically, $\mathcal{L}_{\mathrm{contrast}}$ is an InfoNCE-style contrastive loss applied to cell-level augmentations: for each $x_i$, two stochastic views $x_i^{(1)},x_i^{(2)}$ are produced via Gaussian noise and dropout masking, with corresponding latents $z_i^{(1)},z_i^{(2)}$ \cite{oord2018representation,chen2020simple}:
\begin{equation}
\mathcal{L}_{\mathrm{contrast}}
= -\frac{1}{n}\sum_{i=1}^{n}
\log\frac{\exp\!\left(\mathrm{sim}(z_i^{(1)},z_i^{(2)})/\tau\right)}{\sum_{j=1}^{n}\exp\!\left(\mathrm{sim}(z_i^{(1)},z_j^{(2)})/\tau\right)},
\end{equation}
where $\mathrm{sim}(\cdot,\cdot)$ is cosine similarity and $\tau$ is a temperature parameter. This term preserves local neighborhood structure while reconstruction prevents representation collapse. The VAE uses a probabilistic latent-variable formulation~\cite{kingma2013auto}:
\begin{equation}
\mathcal{L}_{\mathrm{VAE}}
=
-\mathbb{E}_{q_{\theta}(z\mid x)}\!\left[\log p_{\phi}(x\mid z)\right]
+ \beta D_{\mathrm{KL}}\!\left(q_{\theta}(z\mid x)\,\|\,p(z)\right),
\end{equation}
where $\beta$ controls latent-distribution regularization.

\subsection{Projection, Clustering Heads, and Oracle-$k$}
Selected pipelines apply UMAP~\cite{mcinnes2018umap} as a projection $U_{\psi}:Z\rightarrow U$. Clustering uses KMeans, GMM, HDBSCAN, or Leiden. For KMeans and GMM, oracle-$k$ sets the number of clusters to the number of reference cell types: this isolates representation quality from model-selection error but is not fully label-free, and we report the limitation explicitly.

\subsection{Evaluation, Optimization, and Sensitivity}
External evaluation uses Adjusted Rand Index (ARI)~\cite{hubert1985comparing}, Adjusted Mutual Information (AMI)~\cite{vinh2010information}, and Normalized Mutual Information (NMI)~\cite{strehl2002cluster}. Internal diagnostics (silhouette, noise ratio, runtime) help identify failure modes but do not certify biological correctness. Hyperparameters $\eta$ are searched on $\Omega$ via Optuna~\cite{akiba2019optuna} as $\eta^*=\argmax_{\eta\in\Omega}J(\eta)$; robustness is summarized by the mean of repeated runs. Sobol total-order indices~\cite{sobol1990sensitivity,saltelli2010variance} quantify how strongly each hyperparameter contributes to score variance, including interactions.

\section{Experimental Design}
\subsection{Datasets}
\begin{figure}[H]
\centering
\begin{minipage}[t]{0.325\textwidth}
\centering
\figplaceholder{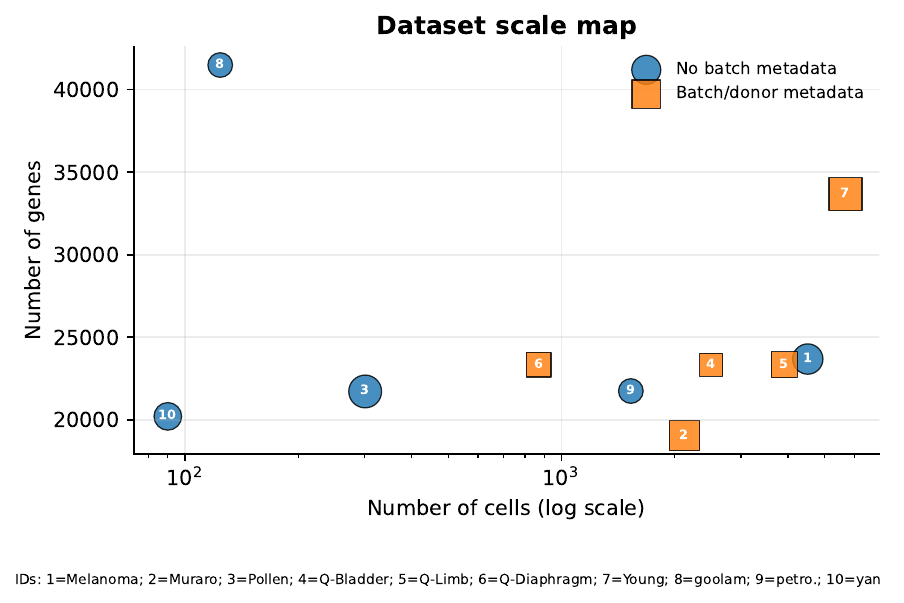}{\textwidth}\\[-0.50em]
{\scriptsize (a) Cell--gene scale map}
\end{minipage}\hfill
\begin{minipage}[t]{0.325\textwidth}
\centering
\figplaceholder{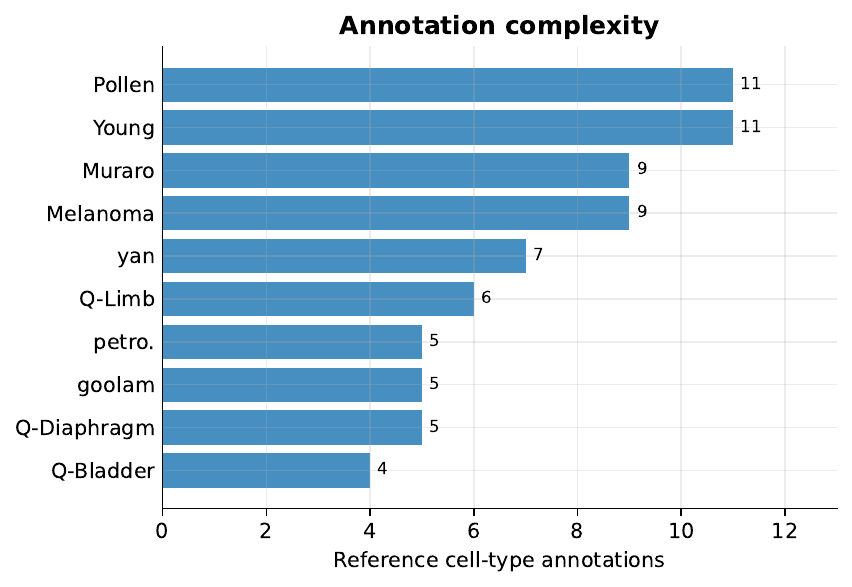}{\textwidth}\\[-0.50em]
{\scriptsize (b) Reference cell types}
\end{minipage}\hfill
\begin{minipage}[t]{0.325\textwidth}
\centering
\figplaceholder{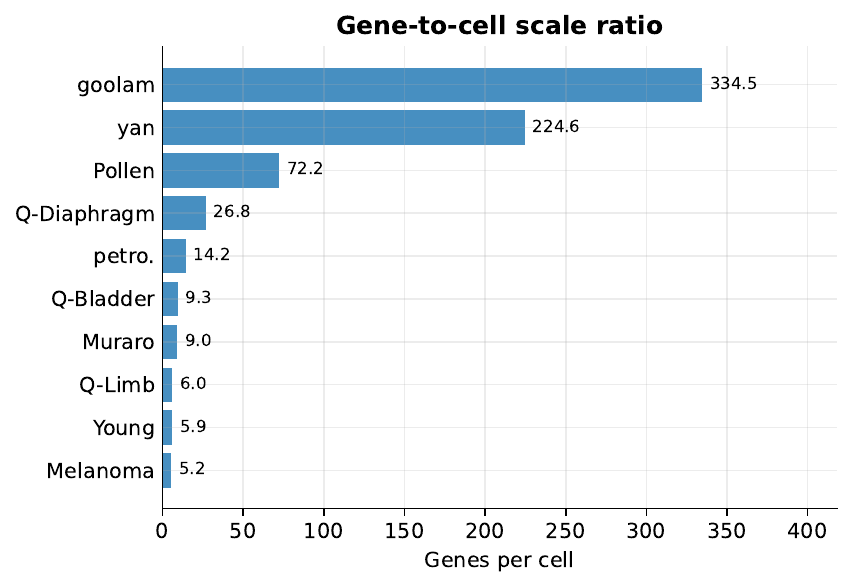}{\textwidth}\\[-0.50em]
{\scriptsize (c) Gene-to-cell ratio}
\end{minipage}

\vspace{0.18em}

\begin{minipage}[t]{0.94\textwidth}
\centering
\figplaceholder{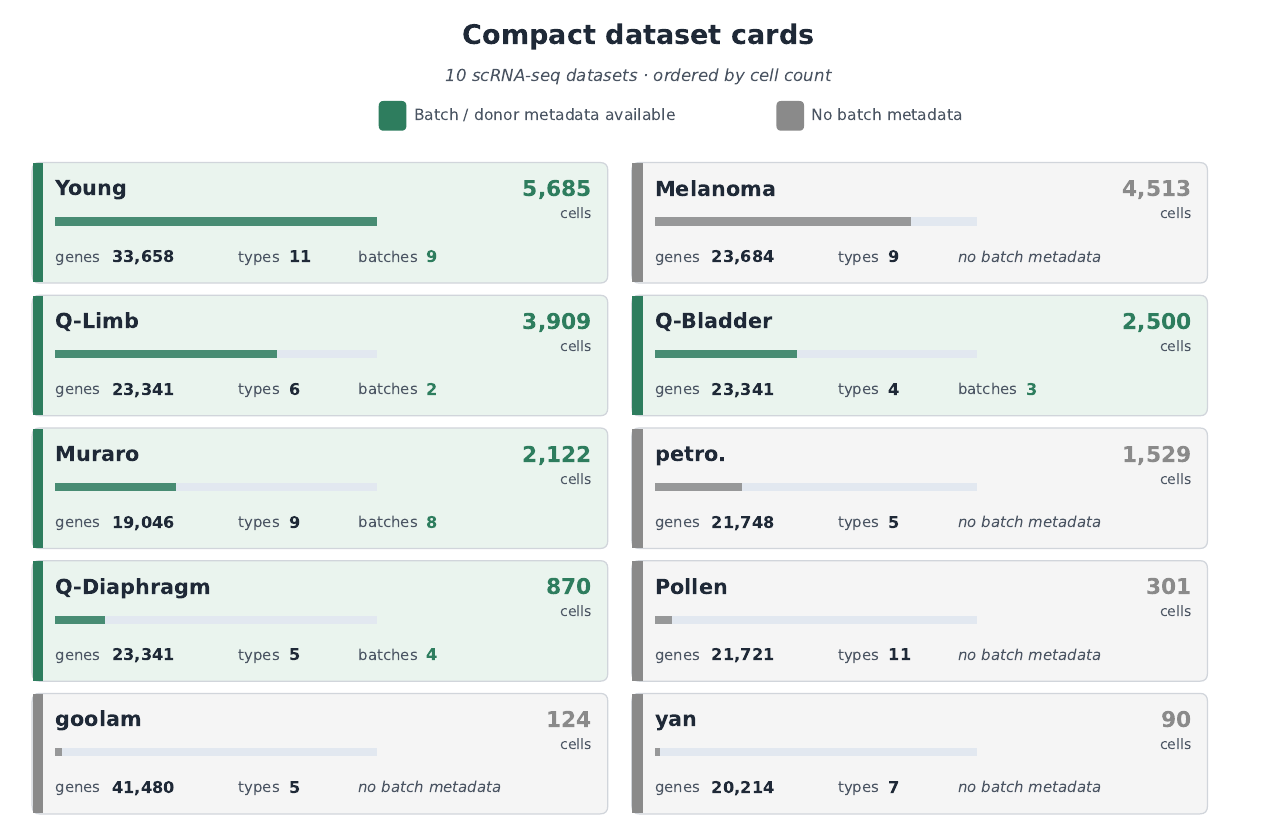}{\textwidth}\\[-0.50em]
{\scriptsize (d) Compact dataset cards}
\end{minipage}

\caption{Dataset profiling. Four views characterize the benchmark before model evaluation: (a)~joint cell--gene scale, (b)~reference cell-type counts, (c)~gene-to-cell ratio (a proxy for sparsity per cell), and (d)~compact metadata cards including batch availability.}
\label{fig:data_profile}
\end{figure}

As summarized in Table~\ref{tab:dataset_metadata}, this study uses ten real scRNA-seq datasets selected to represent heterogeneous experimental regimes, spanning 90--5,685 cells, 19,046--41,480 genes, 4--11 reference annotations, and batch or donor metadata for five datasets. Figure~\ref{fig:data_profile} summarizes the benchmark composition; this profiling view characterizes the data rather than reporting algorithmic performance.

\begin{table}[!htbp]
\centering
\caption{Raw metadata summary of the ten benchmark datasets.}
\label{tab:dataset_metadata}
\scriptsize
\setlength{\tabcolsep}{4pt}
\begin{adjustbox}{width=\textwidth}
\begin{tabular}{lrrrrll}
\toprule
Dataset & Cells & Genes & Types & Batches & Label key & Batch key \\
\midrule
Melanoma\_5K & 4513 & 23684 & 9 & -- & Y & -- \\
Muraro & 2122 & 19046 & 9 & 8 & obs/cell\_ontology\_class & obs/batch \\
Pollen & 301 & 21721 & 11 & -- & Y & -- \\
Quake\_10x\_Bladder & 2500 & 23341 & 4 & 3 & obs/cell\_ontology\_class & obs/donor \\
Quake\_10x\_Limb\_Muscle & 3909 & 23341 & 6 & 2 & obs/cell\_ontology\_class & obs/donor \\
Quake\_Smart\_seq2\_Diaphragm & 870 & 23341 & 5 & 4 & obs/cell\_ontology\_class & obs/donor \\
Young & 5685 & 33658 & 11 & 9 & obs/cell\_ontology\_class & obs/donor \\
goolam & 124 & 41480 & 5 & -- & Y & -- \\
petropoulos & 1529 & 21748 & 5 & -- & Y & -- \\
yan & 90 & 20214 & 7 & -- & Y & -- \\
\bottomrule
\end{tabular}
\end{adjustbox}
\end{table}

\subsection{Compared Methods and Workflow}
\label{sec:methods_workflow}
\begin{figure}[!ht]
\vspace{0.2cm}
\centering
\resizebox{\textwidth}{!}{%
\begin{tikzpicture}[
  font=\sffamily\footnotesize,
  every node/.style={align=center},
  block/.style={
    rectangle, rounded corners=3pt, draw, line width=0.6pt,
    minimum height=12mm, minimum width=26mm,
    text width=26mm, inner sep=3pt,
    drop shadow={shadow xshift=0.5pt, shadow yshift=-0.5pt, opacity=0.25}
  },
  data/.style={block, fill=tierData,    draw=borderData!85},
  rep/.style={block,  fill=tierRep,     draw=borderRep!85},
  repHL/.style={block, fill=tierRep!120, draw=borderRep, line width=0.9pt},
  eval/.style={block, fill=tierEval,    draw=borderEval!85},
  meta/.style={
    rectangle, rounded corners=2.5pt, draw=borderMeta, line width=0.4pt,
    fill=tierMeta, minimum height=10mm, minimum width=28mm,
    text width=28mm, inner sep=2.5pt
  },
  arr/.style={-{Stealth[length=2.6mm,width=2.0mm]}, line width=0.55pt, color=black!65},
  arrFan/.style={-{Stealth[length=2.2mm,width=1.6mm]}, line width=0.45pt, color=black!50},
  arrFB/.style={-{Stealth[length=2.0mm,width=1.4mm]}, line width=0.45pt, dashed, color=borderRep!70},
  tierBadge/.style={
    rectangle, rounded corners=1.5pt, fill=#1, draw=#1!80!black,
    line width=0.3pt, inner sep=2pt, minimum width=20mm,
    font=\sffamily\scriptsize\bfseries, text=white
  },
]

\node[data]                          (raw)   {\textbf{Raw HDF5}\\[-0.5pt]\scriptsize cell\,$\times$\,gene matrix};
\node[data, right=4mm of raw]        (qc)    {\textbf{QC + HVG}\\[-0.5pt]\scriptsize filter, normalize};
\node[data, right=4mm of qc]         (batch) {\textbf{Batch step}\\[-0.5pt]\scriptsize Harmony when available};

\node[tierBadge=borderData, left=4mm of raw, anchor=east] (badgeData) {\,1\;\;DATA\,};

\node[rep,   below=10mm of raw,   xshift=0mm]  (pca)  {\textbf{Classical}\\[-0.5pt]\scriptsize PCA};
\node[repHL, below=10mm of qc]                 (deep) {\textbf{Proposed}\\[-0.5pt]\scriptsize AE\;/\;CAE\;/\;VAE\\[-1pt]{\scriptsize\itshape\color{borderRep!90}\,Optuna-tuned\,$\circlearrowleft$}};
\node[rep,   below=10mm of batch]              (scvi) {\textbf{Specialized}\\[-0.5pt]\scriptsize scVI\,V2};

\node[tierBadge=borderRep, left=4mm of pca, anchor=east] (badgeRep) {\,2\;\;REPRESENTATION\,};

\node[eval, below=10mm of pca]   (umap)    {\textbf{UMAP}\\[-0.5pt]\scriptsize optional projection};
\node[eval, below=10mm of deep]  (clust)   {\textbf{Clustering}\\[-0.5pt]\scriptsize KMeans\,/\,GMM\\[-1.5pt]\scriptsize HDBSCAN\,/\,Leiden};
\node[eval, below=10mm of scvi]  (metrics) {\textbf{Metrics}\\[-0.5pt]\scriptsize ARI\,/\,NMI\,/\,AMI};

\node[tierBadge=borderEval, left=4mm of umap, anchor=east] (badgeEval) {\,3\;\;EVALUATION\,};

\begin{scope}[on background layer]
  \node[
    rectangle, rounded corners=3pt,
    fill=tierMeta!60, draw=borderMeta!60,
    line width=0.4pt, dash pattern=on 1.2pt off 1.0pt,
    fit=(umap)(metrics),
    inner sep=0pt
  ] (dummyfit) {};
\end{scope}

\node[meta, below=14mm of umap]    (optuna) {\textbf{Optuna}\\[-0.5pt]\scriptsize matched budget};
\node[meta, right=3mm of optuna]   (robust) {\textbf{Robustness}\\[-0.5pt]\scriptsize repeated seeds};
\node[meta, right=3mm of robust]   (stats)  {\textbf{Statistics}\\[-0.5pt]\scriptsize Friedman\,/\,Holm\,/\,TOST};
\node[meta, right=3mm of stats]    (sobol)  {\textbf{Sensitivity}\\[-0.5pt]\scriptsize Sobol $S_T$};

\begin{scope}[on background layer]
  \node[
    rectangle, rounded corners=4pt,
    fill=tierMeta, draw=borderMeta,
    line width=0.5pt, dash pattern=on 2pt off 1.5pt,
    fit=(optuna)(sobol),
    inner sep=4pt
  ] (metaPanel) {};
\end{scope}

\node[tierBadge=borderMeta, left=4mm of optuna, anchor=east] (badgeMeta) {\,VALIDATION\,};

\draw[arr] (raw)   -- (qc);
\draw[arr] (qc)    -- (batch);

\draw[arrFan] (batch.south) .. controls +(0,-3mm) and +(0,5mm) .. (pca.north);
\draw[arrFan] (batch.south) -- (deep.north);
\draw[arrFan] (batch.south) .. controls +(0,-3mm) and +(0,5mm) .. (scvi.north);

\draw[arr] (pca)   -- (umap);
\draw[arr] (deep)  -- (clust);
\draw[arr] (scvi)  -- (metrics);

\draw[arr] (umap)  -- (clust);
\draw[arr] (clust) -- (metrics);

\node[
  rectangle, rounded corners=1.5pt,
  fill=white, draw=black!25, line width=0.35pt,
  below=5mm of metaPanel, font=\sffamily\scriptsize,
  text width=110mm, align=center, inner sep=3pt
] (note) {%
  Reference labels are used \emph{only} for evaluation and explicit oracle-$k$ heads;\\[-1pt]
  they never train the representation models.%
};

\end{tikzpicture}%
}
\caption{Three-tier benchmark architecture. \textbf{Tier~1 (Data)} ingests raw HDF5 matrices and applies QC, highly variable gene selection, and optional Harmony-style batch correction. \textbf{Tier~2 (Representation)} compares three families head-to-head: classical PCA, the proposed deep autoencoder variants (highlighted, Optuna-tuned), and specialized scVI~V2. \textbf{Tier~3 (Evaluation)} applies clustering heads and computes external metrics. The dashed \textsc{Validation} panel runs Optuna search, repeated-run robustness, statistical testing, and Sobol sensitivity analysis on the proposed deep representation workflow.}
\label{fig:pipeline}
\end{figure}
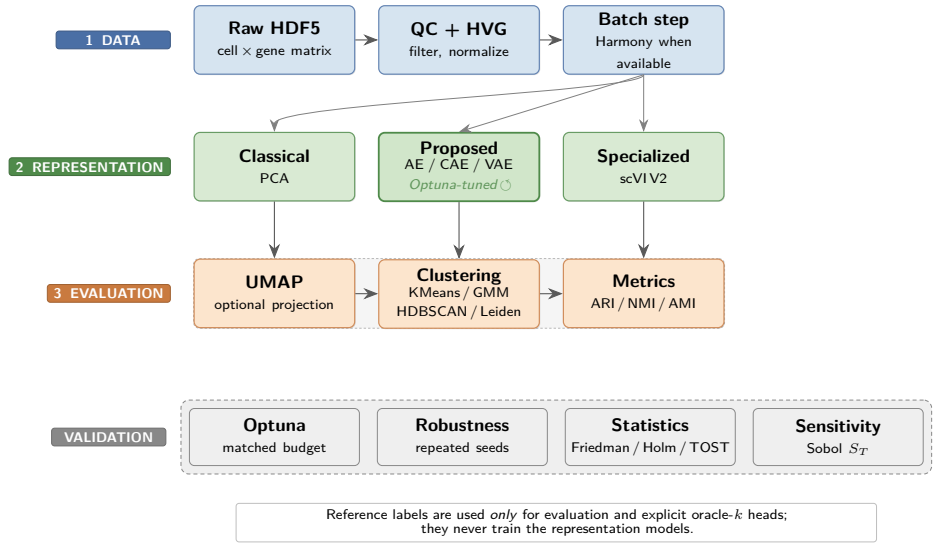

As summarized in Table~\ref{tab:methods}, the benchmark comprises six classical baselines, three proposed deep variants, and three scVI~V2 baselines. Figure~\ref{fig:pipeline} visualizes the workflow as three sequential tiers (data, representation, evaluation) with a meta-layer for validation analyses.

\begin{table}[!htbp]
\vspace{-0.2cm}
\centering
\caption{Methods evaluated in the benchmark.}
\label{tab:methods}
\scriptsize
\setlength{\tabcolsep}{4pt}
\begin{adjustbox}{width=\textwidth}
\begin{tabular}{lll}
\toprule
Group & Method & Description \\
\midrule
Classical & B0\_PCA\_UMAP\_HDBSCAN & PCA, UMAP, HDBSCAN \\
Classical & B1\_PCA\_HDBSCAN & PCA and HDBSCAN \\
Classical & B2\_PCA\_KMEANS & PCA and KMeans with oracle $k$ \\
Classical & B3\_PCA\_GMM & PCA and GMM with oracle $k$ \\
Classical & B4\_PCA\_UMAP\_KMEANS & PCA, UMAP, KMeans with oracle $k$ \\
Classical & B5\_PCA\_UMAP\_LEIDEN & PCA, UMAP, Leiden \\
Proposed & DeepAE--UH & Autoencoder with UMAP--HDBSCAN head \\
Proposed & C-DeepAE--UH & Contrastive autoencoder with UMAP--HDBSCAN head \\
Proposed & VAE--UH & Variational autoencoder with UMAP--HDBSCAN head \\
Specialized & SCVI\_V2\_KMEANS\_K & scVI latent, KMeans with oracle $k$ \\
Specialized & SCVI\_V2\_GMM\_K & scVI latent, GMM with oracle $k$ \\
Specialized & SCVI\_V2\_LEIDEN\_KMATCH & scVI latent, Leiden resolution matched to cluster count \\
\bottomrule
\end{tabular}
\end{adjustbox}
\end{table}

\paragraph{Coverage of the scVI~V2 baseline.}
The scVI~V2 pipeline was initially incomplete on three datasets in our protocol, namely Muraro, Pollen, and yan, where stable embeddings were not produced within the matched compute budget. For yan ($90$ cells) and Pollen ($301$ cells), the very small cell counts and high gene-to-cell ratios ($224.6$ and $72.2$) fall outside the typical data regime for which scVI was originally designed, which is consistent with the difficulty observed. The case of Muraro ($2122$ cells, $8$ batches) is less obvious from raw scale alone and may reflect interactions between batch handling, hyperparameter defaults, and the matched training budget used for fairness across methods. We therefore report scVI on the seven-dataset intersection and treat the comparison as partial; extending it to all ten datasets, including a careful re-tuning on Muraro, is a natural direction for future work.

\section{Results}
\subsection{Main Benchmark and scVI V2 Comparison}
Table~\ref{tab:main_mean_rank} summarizes the original nine-method benchmark across ten datasets. \mbox{C-DeepAE--UH} achieved the highest mean ARI ($0.7872$), with PCA--KMeans, PCA--GMM, and PCA--UMAP--KMeans remaining competitive. Figure~\ref{fig:meanrank} extends the comparison to all twelve methods including the partial scVI~V2 baselines.

\begin{table}[!htbp]
\centering
\caption{Original nine-method benchmark across ten datasets. Lower mean rank is better.}
\label{tab:main_mean_rank}
\scriptsize
\begin{adjustbox}{width=0.78\textwidth}
\begin{tabular}{lrrrr}
\toprule
Method & Mean ARI & Mean rank & Wins & Datasets \\
\midrule
C-DeepAE--UH & 0.7872 & 3.65 & 1.5 & 10 \\
DeepAE--UH & 0.7567 & 3.65 & 1.5 & 10 \\
B2 PCA-KMeans & 0.7433 & 3.90 & 2.0 & 10 \\
VAE--UH & 0.7362 & 4.10 & 1.0 & 10 \\
B4 PCA-UMAP-KMeans & 0.7448 & 4.30 & 1.0 & 10 \\
B0 PCA-UMAP-HDBSCAN & 0.7580 & 4.45 & 0.5 & 10 \\
B3 PCA-GMM & 0.7574 & 5.00 & 2.0 & 10 \\
B5 PCA-UMAP-Leiden & 0.4299 & 7.55 & 0.5 & 10 \\
B1 PCA-HDBSCAN & 0.4433 & 8.40 & 0.0 & 10 \\
\bottomrule
\end{tabular}
\end{adjustbox}
\end{table}

\begin{center}
\vspace{-0.2cm}
\begin{minipage}{0.94\textwidth}
\centering
\figplaceholder{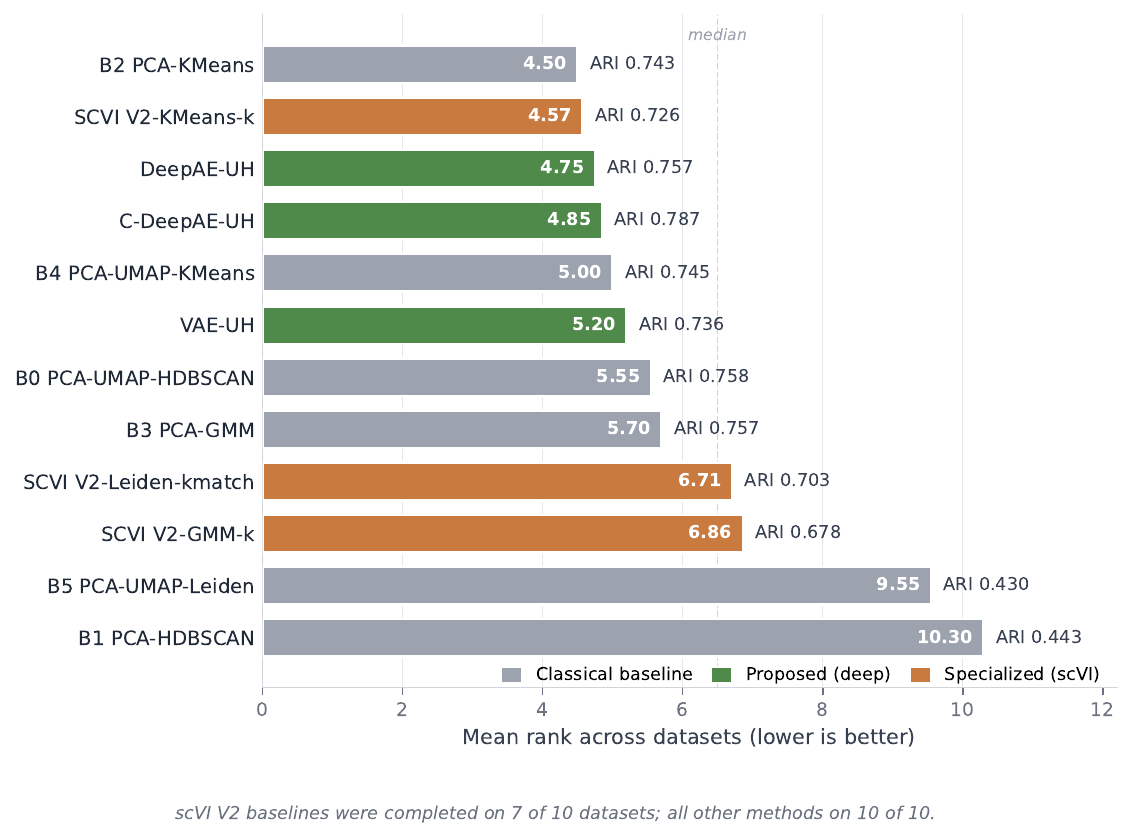}{0.80\textwidth}
\captionof{figure}{Mean-rank comparison of all twelve clustering pipelines. Lower rank indicates better average position across datasets. The scVI~V2 baselines were completed on seven of ten datasets.}
\label{fig:meanrank}
\end{minipage}
\end{center}

Table~\ref{tab:scvi_intersection} summarizes the seven-dataset intersection between the proposed methods and scVI~V2 variants. SCVI V2-KMeans-k achieved the best mean rank, whereas \mbox{VAE--UH} and \mbox{C-DeepAE--UH} achieved higher mean ARI. The divergence between rank and mean ARI arises because rank rewards consistent per-dataset ordering whereas mean ARI is sensitive to the magnitude of individual gaps; together they indicate dataset-dependent performance rather than dominance by any one family.

\begin{table}[!htbp]
\centering
\caption{Intersection comparison between proposed and scVI~V2 variants on seven datasets.}
\label{tab:scvi_intersection}
\scriptsize
\begin{adjustbox}{width=0.62\textwidth}
\begin{tabular}{lrrr}
\toprule
Method & Mean ARI & Mean rank & Datasets \\
\midrule
SCVI V2-KMeans-k & 0.7262 & 2.43 & 7 \\
DeepAE--UH & 0.7331 & 3.50 & 7 \\
VAE--UH & 0.7654 & 3.57 & 7 \\
SCVI V2-Leiden-kmatch & 0.7026 & 3.71 & 7 \\
C-DeepAE--UH & 0.7522 & 3.79 & 7 \\
SCVI V2-GMM-k & 0.6783 & 4.00 & 7 \\
\bottomrule
\end{tabular}
\end{adjustbox}
\end{table}

\subsection{Statistical Testing}
The Friedman test on the complete twelve-method, seven-dataset paired subset indicated a significant overall method effect ($\chi^2{=}33.4845$, $p{=}4.39{\times}10^{-4}$). Holm-corrected Wilcoxon tests, however, did not establish significant superiority of the proposed variants over the strongest classical or scVI baselines, and TOST tests with a strict $0.02$ ARI margin did not establish equivalence either. The $0.02$ ARI margin was chosen conservatively because it is comparable to the repeated-run variability observed across baseline methods ($0.01$--$0.03$ ARI), making differences at this scale unlikely to affect downstream cell-population interpretation. This means the data are consistent with method-dependent variation in mean ranks but do not support a sharp claim of dominance; this is precisely the regime where dataset-aware analysis becomes informative. Figure~\ref{fig:perdataset} shows that the per-dataset winners scatter across classical, proposed, and specialized families, and Figure~\ref{fig:stat} summarizes mean ARI gaps relative to the contrastive variant.

\begin{center}
\begin{minipage}{0.72\textwidth}
\centering
\figplaceholder{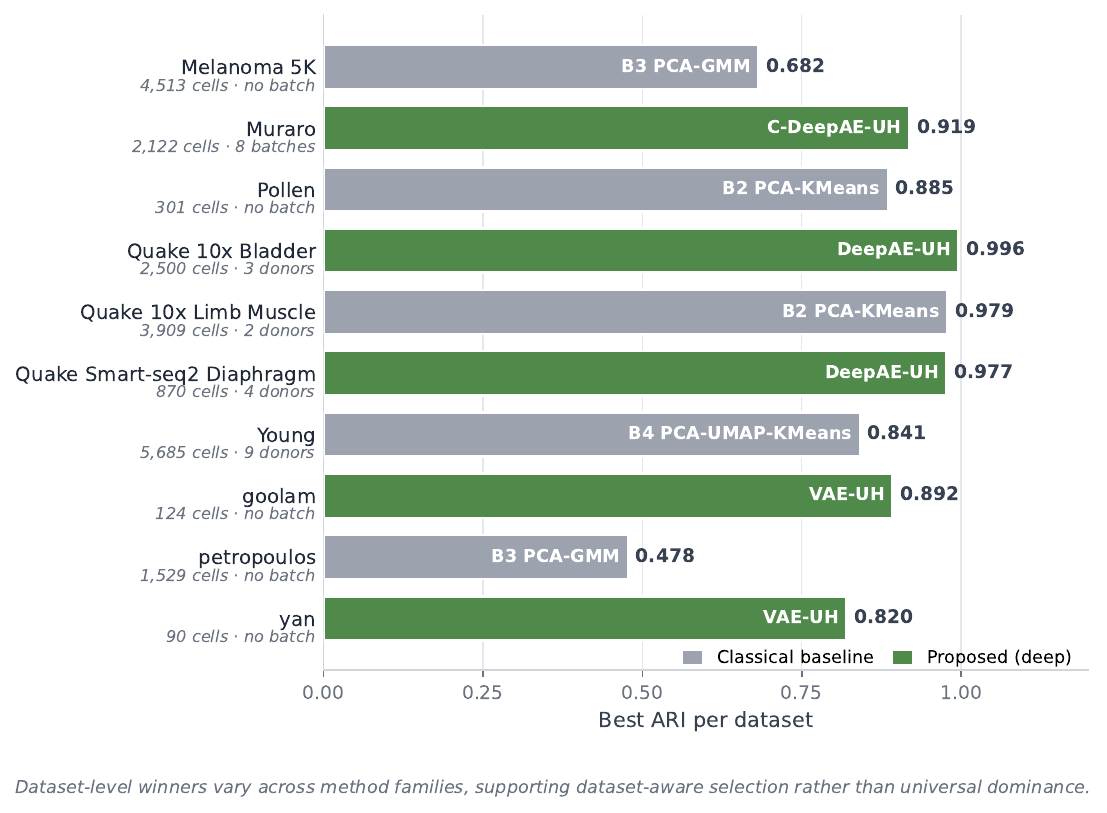}{0.92\textwidth}
\captionof{figure}{Per-dataset winning method and best ARI across the ten benchmark datasets.}
\label{fig:perdataset}
\end{minipage}
\end{center}

\begin{center}
\begin{minipage}{0.90\textwidth}
\centering
\figplaceholder{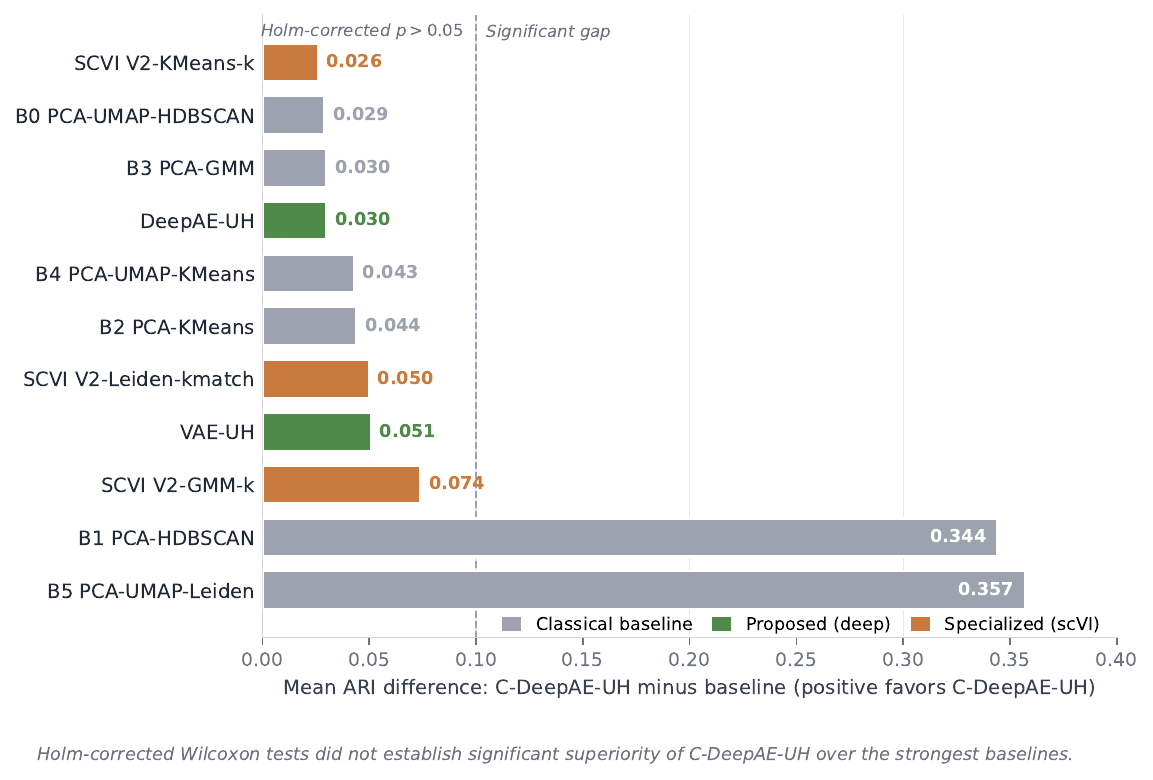}{0.82\textwidth}
\captionof{figure}{Mean ARI differences between \mbox{C-DeepAE--UH} and representative baselines (positive favors C-DeepAE--UH). The dashed line at $0.10$ separates baselines that are competitive after Holm correction from the two clearly weaker baselines (B1, B5).}
\label{fig:stat}
\end{minipage}
\end{center}

\subsection{Three Dataset Regimes}
\label{sec:regimes}
Cross-referencing Figure~\ref{fig:perdataset} with the metadata in Table~\ref{tab:dataset_metadata} reveals three reproducible regimes that explain the per-dataset outcomes and motivate dataset-aware method selection. The pattern is not a clean split between deep and classical methods; it reflects how dataset properties such as cell count, annotation complexity, and batch metadata interact with each representation family.

\paragraph{Regime A: very small datasets with sparse-per-cell signal.}
On yan ($90$ cells, ratio~$224.6$) and goolam ($124$ cells, ratio~$334.5$), \mbox{VAE--UH} achieved the best ARI ($0.82$ and $0.89$ respectively). When the sample is far smaller than the feature dimension, the KL term in the VAE objective acts as an effective inductive bias, smoothing the latent space and protecting against overfitting on dominant noise modes. We note that Pollen ($301$ cells, $11$ types) is also small but is best served by classical KMeans rather than VAE, suggesting that sample size alone is not sufficient: the high annotation complexity of Pollen rewards a method that recovers many compact clusters in a linear subspace.

\paragraph{Regime B: mid-scale datasets where deep variants extract additional structure.}
On Quake\_10x\_Bladder ($2{,}500$ cells, $3$ donors) and Quake\_Smart\_seq2\_Diaphragm ($870$ cells, $4$ donors), the vanilla \mbox{DeepAE--UH} achieved the highest ARI ($0.996$ and $0.977$). Muraro ($2{,}122$ cells, $8$ batches, $9$ types) is best served by \mbox{C-DeepAE--UH} ($0.919$): with many batches and many fine-grained cell types, the contrastive objective directly targets neighborhood preservation across batch-driven shifts, and the autoencoder reconstruction prevents collapse onto batch-specific modes.

\paragraph{Regime C: datasets where classical PCA pipelines remain competitive.}
On Quake\_10x\_Limb\_Muscle, Young, and Melanoma, classical PCA-based pipelines (B2 PCA-KMeans, B4 PCA-UMAP-KMeans, B3 PCA-GMM) match or exceed the deep variants. The common factor is that the dominant variation in these datasets is well captured by linear projection, so the additional non-linear capacity of an autoencoder yields no measurable benefit and a simpler model is preferable on compute grounds. The petropoulos dataset is an outlier: all methods score below $0.50$, indicating that none of the evaluated representations resolves its annotation structure well, regardless of family.

\subsection{Sobol Sensitivity}
Sobol total-order indices in the deep representation workflow rank hyperparameters by their contribution to score variance: learning rate ($S_T{=}0.6972$), latent dimensionality ($S_T{=}0.5604$), training budget ($S_T{=}0.5024$), dropout ($S_T{=}0.4783$), UMAP minimum distance ($S_T{=}0.4184$), and UMAP neighborhood size ($S_T{=}0.2892$). Figure~\ref{fig:sobol} visualizes these values. The practical implication is direct: a fixed tuning budget should prioritize learning rate and latent size before sweeping UMAP graph parameters. Reallocating compute from low-impact to high-impact knobs reduces the number of trials required to reach a given ARI level, which is the concrete sense in which the analysis ties methodological choice to compute efficiency.

\begin{center}
\begin{minipage}{0.88\textwidth}
\centering
\figplaceholder{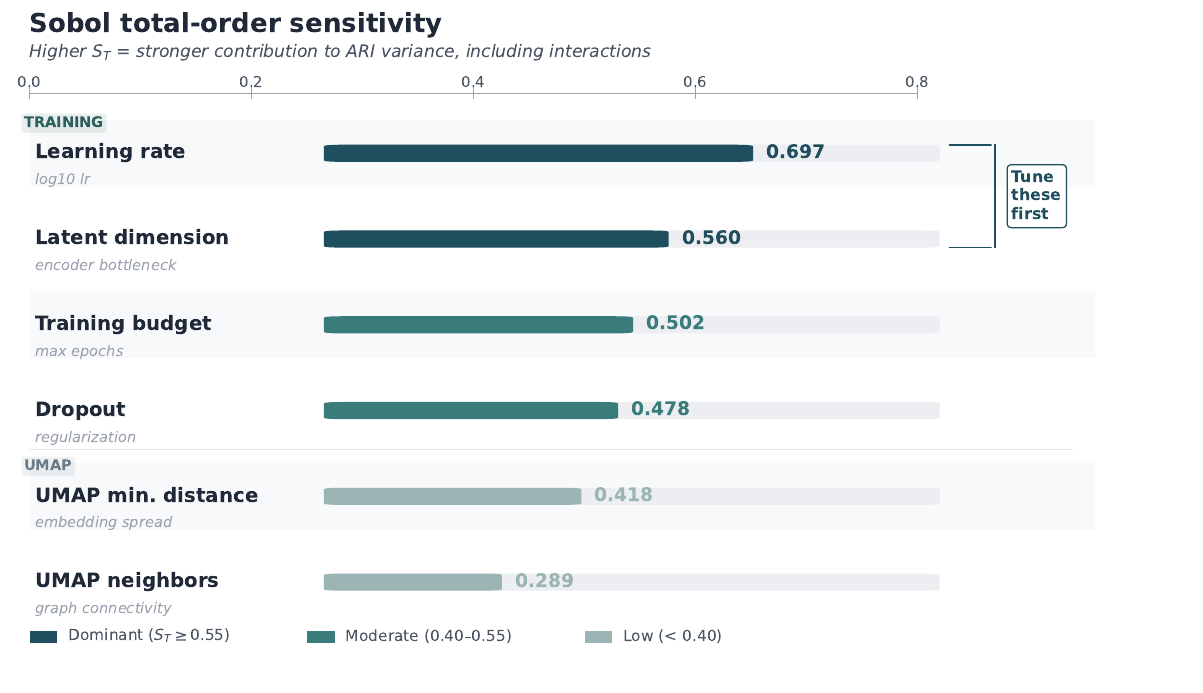}{0.82\textwidth}
\captionof{figure}{Sobol total-order sensitivity of the deep representation workflow. Higher values indicate stronger contribution to ARI variance, including interactions.}
\label{fig:sobol}
\end{minipage}
\end{center}

\section{Discussion and Limitations}
\paragraph{What the benchmark answers.}
The motivating question of when an additional deep representation stage measurably helps has a defensible answer, just not a universal one. Probabilistic regularization (VAE) helps on the smallest datasets where deterministic encoders overfit (Regime~A); deep autoencoders, including the contrastive variant, help on mid-scale data where multi-batch or many-type structure rewards explicit neighborhood preservation (Regime~B); classical PCA pipelines remain competitive at substantially lower compute cost when the dominant variation is already linear (Regime~C). The defensible recommendation is therefore dataset-aware selection rather than blanket adoption of a single architecture.

\paragraph{What the sensitivity analysis adds.}
Sobol total-order indices localize where the hyperparameter budget should be spent: learning rate ($S_T{=}0.70$) and latent dimensionality ($S_T{=}0.56$) dominate, while UMAP neighborhood size contributes the least ($S_T{=}0.29$). Practitioners with a fixed compute budget can therefore freeze low-impact knobs at reasonable defaults and concentrate search on the high-impact ones. This is the operational consequence of the analysis and the explicit link between methodological choice and compute efficiency in biomedical pipelines.

\paragraph{Why this matters beyond the benchmark.}
Two takeaways generalize. First, conservative statistical reporting, namely Friedman with Holm correction and TOST equivalence, prevents the common practice of selectively highlighting raw ARI gaps that do not survive multiple-comparison correction. Second, sensitivity analysis turns a methodological study into a budget-aware deployment guide, which is what practical biomedical AI pipelines need.

\paragraph{Limitations.}
(i)~Reference annotations serve as evaluation labels but are not absolute biological truth, because cell-type labels can be incomplete, noisy, or inaccurate in practical single-cell integration and evaluation settings~\cite{andreatta2024semi}. (ii)~The scVI comparison is partial (seven of ten datasets), as discussed in Section~\ref{sec:methods_workflow}. (iii)~Oracle-$k$ supports controlled representation comparison but is not fully label-free, because estimating the number of cell types remains a distinct and nontrivial clustering problem in scRNA-seq analysis~\cite{yu2022benchmarking}. (iv)~Batch-aware correction is possible only for the five datasets with batch metadata. (v)~Recent graph-contrastive, self-supervised, transfer-learning, and foundation-model approaches for single-cell analysis were not fully reproduced~\cite{cui2024scgpt}. (vi)~The Sobol analysis was performed on a representative configuration; cross-dataset and cross-method extension, together with evaluation on large-scale contemporary atlases such as Tabula Sapiens and CELLxGENE corpora under matched compute budget, is left for future work.

\section{Conclusion}
We presented a diagnostic benchmark of deep representation learning for scRNA-seq clustering on ten real datasets. Although autoencoder, contrastive, and variational variants are competitive with strong classical baselines and a partial scVI comparison, Holm-corrected tests do not establish universal superiority. Per-dataset analysis instead exposes three reproducible regimes (VAE on the smallest datasets, deep autoencoders on mid-scale multi-batch or many-type data, classical PCA pipelines when linear projection captures the dominant variation), and Sobol sensitivity ranks learning rate and latent dimensionality as the dominant variance contributors. The result is a dataset-aware, compute-conscious workflow for biomedical single-cell data mining; extension to graph-contrastive and self-supervised models is left for future work.

\begin{credits}
\subsubsection{Acknowledgement}
This research is funded by Van Lang University, Vietnam under grant number~2510-DT-VLT-KCT-SV-002. The authors gratefully acknowledge Van Lang University for its financial and academic support during the implementation of this study.
\end{credits}

\bibliographystyle{IEEEtran}
\bibliography{IEEEabrv,reference}
\end{document}